\documentclass{article}

\usepackage{arxiv}

\usepackage{cite}
\usepackage{amsmath,amssymb,amsfonts}
\usepackage{algorithmic}
\usepackage{graphicx}
\usepackage{textcomp}

\usepackage{longtable}

\usepackage{booktabs}

\usepackage{lineno}

\usepackage{hyperref}

\usepackage{multirow}
\usepackage{subcaption}

\usepackage{bm}

\def\BibTeX{{\rm B\kern-.05em{\sc i\kern-.025em b}\kern-.08em
    T\kern-.1667em\lower.7ex\hbox{E}\kern-.125emX}}

\title{A Machine Learning Approach to Automatic Fall Detection of Soldiers}

\author{
 Leandro Soares \\
  Cefet/RJ, IPqM - Brazilian Navy\\
  Rio de Janeiro, Brazil \\
  \texttt{leandro.soares@aluno.cefet-rj.br} \\
   \And
 Gustavo Venturini \\
  IPqM - Brazilian Navy \\
  Rio de Janeiro, Brazil \\
  \texttt{corteletti@marinha.mil.br} \\
  \And
 José Gomes \\
  IPqM - Brazilian Navy \\
  Rio de Janeiro, Brazil \\
  \texttt{junior.carvalho@marinha.mil.br} \\
  \AND
  Jonathan Efigenio \\
  IPqM - Brazilian Navy \\
  Rio de Janeiro, Brazil \\
  \texttt{jonathan.efigenio@marinha.mil.br} \\
  \And
  Pablo Rangel \\
  IPqM - Brazilian Navy \\
  Rio de Janeiro, Brazil \\
  \texttt{pablo.rangel@marinha.mil.br} \\
  \And
  Pedro Henrique Gonzalez \\
  UFRJ \\
  Rio de Janeiro, Brazil \\
  \texttt{pegonzalez@cos.ufrj.br} \\
  \And
  Joel dos Santos \\
  Cefet/RJ \\
  Rio de Janeiro, Brazil \\
  \texttt{joel.santos@cefet-rj.br} \\
  \And
  Diego Brandão \\
  Cefet/RJ \\
  Rio de Janeiro, Brazil \\
  \texttt{diego.brandao@cefet-rj.br} \\
  \And
  Eduardo Bezerra \\
  Cefet/RJ \\
  Rio de Janeiro, Brazil \\
  \texttt{ebezerra@cefet-rj.br} \\
}

\begin{document}
\maketitle

\begin{abstract}
Military personnel and security agents often face significant physical risks during conflict and engagement situations, particularly in urban operations. Ensuring the rapid and accurate communication of incidents involving injuries is crucial for the timely execution of rescue operations. This article presents research conducted under the scope of the Brazilian Navy's ``Soldier of the Future'' project, focusing on the development of a Casualty Detection System to identify injuries that could incapacitate a soldier and lead to severe blood loss. The study specifically addresses the detection of soldier falls, which may indicate critical injuries such as hypovolemic hemorrhagic shock. To generate the publicly available dataset, we used smartwatches and smartphones as wearable devices to collect inertial data from soldiers during various activities, including simulated falls. The data were used to train 1D Convolutional Neural Networks (CNN1D) with the objective of accurately classifying falls that could result from life-threatening injuries. We explored different sensor placements (on the wrists and near the center of mass) and various approaches to using inertial variables, including linear and angular accelerations. The neural network models were optimized using Bayesian techniques to enhance their performance. The best-performing model and its results, discussed in this article, contribute to the advancement of automated systems for monitoring soldier safety and improving response times in engagement scenarios.
\end{abstract}


\maketitle

\section{Introduction}
\label{introdction}

Military personnel and security agents face physical risks in a variety of conflict and combat scenarios, both in urban and military contexts. In the military context, these scenarios include the retaking of military facilities, hostage rescue, target neutralization, and city defense, as well as United Nations peacekeeping missions. In urban operations, such as those carried out in low-income communities dominated by criminals in Brazil, the \textit{Guarantee of Law and Order} (GLO) operations stand out.

According to the GLO Manual~\cite{MD-GLO}, GLO operations are conducted by the Armed Forces in serious situations of public order disruption when traditional public security forces are exhausted. In Brazil, these emergency operations are determined by the President of the Republic, delimiting a specific area and for a limited period. The goal of these operations is to ensure the preservation of public order, the safety of individuals, and the protection of property. Since 1992, the Brazilian Ministry of Defense has documented 146 GLO operations, 46 involving  Brazilian  Marines~\cite{MD-GLO-2}.

On the global stage, Brazil has played an active role in recent years, participating in over 30 United Nations peacekeeping missions  and sending over 27,000 military personnel abroad for these missions~\cite{PK}. A significant example is the MINUSTAH operation, which took place between 2004 and 2017, aimed at stabilizing Haiti after a civil conflict~\cite{minustah}. In this operation, the Brazilian Navy contributed naval forces and marine troops to support Haiti's security and reconstruction.

Additionally, it is important to mention the incursions of military police into low-income communities and slums in urban areas, which assume guerrilla-like characteristics and are part of daily life in Rio de Janeiro municipality. The Rio de Janeiro Military Police has recorded 477 police deaths in action since 2003~\cite{ISP-DADOS}.

During those mentioned operations, soldiers are exposed to the risk of injuries from confrontations with enemy troops, with the most common being firearm injuries. Such injuries can lead to the soldier's death if immediate communication of the incident with \textit{Command and Control} (C2) is not established, and prompt rescue operations are not carried out. The first ten minutes after the injury are critical, essential for stabilizing the victim, and preventing the worsening of their condition. Procedures performed during this period, known as the ``Platinum 10'', often save lives. Additionally, the first hour following the incident, known as the ``Golden Hour'', is the crucial time in which the victim has the highest chances of survival, provided he/she receives medical care within this period~\cite{AP-HEALTH}.

Given the above considerations, it is vital for maintaining the life of a wounded soldier that the incident information reaches C2 quickly and accurately, allowing for the swift mobilization of the rescue operation. Hence, it becomes necessary to research and develop methods for the immediate detection of a soldier's incapacitation and the swift transmission of this information to the command and control system.

Considering the active participation of its soldiers in military operations that may result in armed conflicts, as well as the operations of other Armed Forces and Military Police, the Brazilian Navy, through IPqM, initiated the project called ``Soldier of the Future''. Part of this project's scope is the study of biometric parameters that can be collected from soldiers in action, aiming at the automatic inference of relevant incidents for support and/or safeguarding procedures.

The research described in this paper is part of the ``Soldier of the Future'' project and aims to implement a Casualty Detection System. This work focuses on one component of this system, which is designed to identify injuries that could incapacitate a soldier and lead to significant blood loss. The system is composed of two models: (1) detection of the soldier's fall based on inertial variables, and (2) classification of the fall as a result of hypovolemic hemorrhagic shock or an operational fall. This paper specifically focuses on the first model.

Due to the nature of military operations, it is necessary for devices used for fall detection to be lightweight, easy to use, and easy to wear. Therefore, for the collection of biometric data and consequent fall detection, smartwatches and smartphones were chosen as portable wearable devices for collecting inertial variables. Data collection for the research involved 15 volunteer soldiers from the Brazilian Navy who performed various activities categorized as daily life activities, armed service activities, and military operations, in addition to simulated fall activities. Each participant wore a smartwatch on each wrist and a smartphone in the pocket of their uniform jacket, allowing for data collection from the right and left wrists as well as from an area near the body's center of mass. This article presents the employed methodology, the obtained results, and provides a public dataset called IPqM-Fall at \url{https://github.com/AILAB-CEFET-RJ/falldetection} to support future research in this field.

Artificial intelligence techniques, especially machine learning techniques such as artificial neural networks, have been widely applied to classification problems~\cite{mukhamediev2022}. In this paper, we conducted fall detection experiments using 1D Convolutional Neural Networks (CNN1D). These networks were trained with data in both the time domain and the frequency domain.

Additionally, the networks were trained separately with data from smartwatches worn on the right wrist, left wrist, and a smartphone positioned on the chest. The goal was to compare which location would be most suitable for sensor portability, considering predictive performance in fall classification. We also conducted experiments using only the magnitude of the linear acceleration of the accelerometer, as well as the $x$, $y$, and $z$ components of this acceleration.

The same procedure was applied to the angular acceleration of the gyroscope, and we experimented with an approach that combined the linear and angular acceleration data. The aim was to determine which of these approaches in using inertial variables would be more efficient. These different approaches were organized into multiples experimental pipelines, each identified by an acronym (see Section~\ref{sec:Data Analysis}).

The neural network parameters for all the mentioned pipelines were optimized using Bayesian techniques.

The results obtained in the ``Future Combatant'' project demonstrated the effectiveness of an approach based on one-dimensional convolutional neural networks (CNN1D) for fall detection in military activities. The model optimized with time-domain data, collected from the combatant's chest and using the $x$, $y$, and $z$ components of linear and angular acceleration, achieved the best performance, with MCC values of 0.9952, sensitivity of 1.0000, and specificity of 0.9914.

This paper is divided into five sections. Section~\ref{introdction} corresponds to the already presented introduction, while Section~\ref{sec:Related:Work} addresses related work, covering relevant research that underpinned this study. Section~\ref{sec:Materials and Methods} is dedicated to materials and methods, describing the IPqM-FALL dataset, data collection methods, data preparation and analysis, the architectures of the neural networks used, and the Bayesian optimization process. Section~\ref{sec:results} presents the results obtained, accompanied by their discussion. Finally, Section~\ref{sec:Conclusions} on the conclusion provides the final considerations on the research conducted.

\section{Related Work}
\label{sec:Related:Work}

Fall detection is a widely explored field of research, aiming to develop effective technologies to prevent such incidents, various approaches have been applied to this issue. Vision-based systems, for instance, use cameras to monitor real-time activities, detecting falls through video analysis. Image processing and computer vision techniques are employed to identify changes in body posture.

Environmental sensors, such as floor pressure sensors and motion sensors in residential environments, are also used to detect falls by analyzing movement patterns in physical space. Additionally, IoT (Internet of Things)-based systems combine data from multiple sensors, expanding the spatial coverage of detection.

Among these approaches, machine learning techniques stand out for their growing use due to their generalization and data classification capabilities~\cite{usmani2021}. Algorithms such as k-Nearest Neighbors (k-NN), Support Vector Machines (SVM), and Decision Trees have been applied to differentiate falls from other activities based on sensor data. Recently, deep learning approaches, such as Convolutional Neural Networks (CNN) and Recurrent Neural Networks (RNN), have demonstrated better results in human fall detection task. These algorithms enable the creation of more accurate and adaptive systems, capable of identifying complex movement patterns and distinguishing falls from other activities.

Since 2014, 161 papers focusing on fall detection using artificial intelligence techniques and wearable devices have been published in the Scopus database. The search was conducted with the terms: (``fall detection'' OR ``fall recognition'') AND (``machine learning'' OR ``artificial intelligence'' OR ``deep learning'') AND (``wearable devices'') on June 5, 2024. On the Web of Science database, 84 papers were found using the same search terms. None of the found papers have applications for fall detection during military operations. When conducting a broader search in both databases using the term: (``fall detection'' OR ``fall recognition'') AND (``soldier'' OR ``combatant'' OR ``military'' OR ``army'' OR ``military personnel'' OR ``navy'') on the same date, 18 papers were found. However, none of the papers found in this search address the detection of falls of soldiers in combat action, pointing to a research gap in fall detection applied to the armed forces. 

Even considering the gap in military applications, many studies on fall detection focusing on the elderly provide valuable contributions and are related to the present work. In the following paragraphs, we summarize these studies.

In~\cite{vavoulas2016mobiact}, the authors conducted the data collection and made available the dataset named MobiAct, a dataset intended for fall recognition. The authors describe the data collection, which involved the participation of 66 volunteers who performed four fall activities and nine daily living activities, using smartphones equipped with sensors such as accelerometers and gyroscopes. In the same work, the authors carried out classification experiments with various machine learning algorithms (IBk, J48, Logistic Regression, Multilayer Perceptron, and LMT), where the best result obtained was an accuracy of 99.88\% with the IBk classifier. The method adopted by the authors for collecting data and organizing the MobiAct dataset served as the primary inspiration and structural example for the dataset proposed in this work.
   
In~\cite{mauldin2018smartfall}, the authors propose a system for fall detection using a Microsoft Band smartwatch and a smartphone. The SmartFall app runs on the smartphone and analyzes accelerometer data from the smartwatch to identify falls in real time. To test the system to be implemented, the authors collected data from seven volunteers, all wearing an MS Band watch on their left hand and a Notch device on their wrist. The volunteers performed a predetermined set of simulated falls (forward falls, backward falls, left side falls, and right side falls) and daily activities (running, sitting, throwing an object, and waving hands), and they published the SmartFall dataset. The authors used SVM, Naive Bayes, and a recurrent neural network (RNN) algorithms in their classification experiments, using the SmartFall and Farseeing databases. The RNN achieved the best offline results for the three datasets: Smartwatch: 0.77 precision, 1.0 sensitivity, 0.85 accuracy; Notch: 0.79 precision, 0.89 sensitivity, 0.99 accuracy; Farseeing: 0.37 precision, 1.0 sensitivity, 0.99 accuracy. The authors' work was essential to help choose the smartwatch as a wearable device to be applied in the Future Soldier project and to select the types of autumn activities to be imposed on the volunteers in the dataset proposed in this work.
    
In~\cite{ozdemir2016analysis}, the author presents a machine learning-based fall detection system. The system employs six three-degree-of-freedom orientation tracking units placed on different parts of the volunteer's body (head, chest, wrist, waist, ankle, and thigh) to assess which part of the body is most effective for sensor use in fall detection. The author employed four distinct classification techniques in their experiments: k-Nearest Neighbor Classifier (k-NN), Bayesian Decision Making (BDM), Support Vector Machines (SVM), and Multilayer Perceptron (MLP). The best result, considering the use of a single sensor, was 99.87\%  of precision and 98.42\% of accuracy for the sensor located at the waist using the k-NN algorithm. The author's study prompted us to collect data from volunteer military personnel not only from the wrist but also from the chest, in order to determine the position at which the sensor should be equipped on soldiers during operations.

In~\cite{georgakopoulos2020change}, the authors utilized the MSB and UP-Fall datasets to test their fall detection approach, which involves combining the cumulative sum algorithm (CUSUM) used for change detection and a convolutional neural network (CNN-shallow, CNN-3B3, CNN-1Conv, and CNN-3Conv) to classify the temporal window corresponding to the change. In addition to CNNs, approaches with other machine learning algorithms (LSTM, LDA, SVM, KNN, and MLP) were tested. The CNN-shallow achieved the best result for both datasets: MSB: AUC of 91.43\%, accuracy of 90\%, sensitivity of 63\%; UP-Fall: AUC of 94.01\%, accuracy of 79\%, sensitivity of 94\%. The authors' work emphasizes the need to use a model with low computational complexity to save wearable device battery life, which is ideal for military operations where the duration may be uncertain, making battery autonomy a relevant concern. 

In~\cite{santos2019accelerometer}, the authors explore the use of deep learning techniques, specifically Convolutional Neural Networks (CNNs), for human fall detection using accelerometer data. The study reviews recent works on fall detection, highlighting the effectiveness of deep learning techniques in achieving high accuracy rates. They proposed the model called CNN-3B3Conv, which is a 1D CNN designed to operate in an Internet of Things (IoT) and cloud computing environment. The experiments conducted show that the proposed CNN model, utilizing data augmentation technique, achieved better results than the other tested models, reaching an MCC of 0.9644, an accuracy of 99.13\%, a precision of 100\%, and a specificity of 93.36\%. The contributions of this article were important for defining the 1D CNN architecture as the architecture to be used in our experiments.

\section{Materials and Methods}
\label{sec:Materials and Methods}

This section is dedicated to detailing the methods employed in data collection, the creation of the IPqM-Fall dataset, as well as data preparation and the selection of neural network architectures. This section is organized as follows. Section~\ref{sec:System Design} explores the operation of the casualty detection system, explaining in detail how the data is monitored and analyzed to identify casualty occurrences. Section~\ref{sec:IPqM Fall Dataset} describes the data collection procedures and the structure of the IPqM-Fall dataset. Section~\ref{sec:Construction of the Fall Detection Neural Network Model} discusses the selection and configuration of neural network architectures, also addressing Bayesian hyperparameter optimization and the different testing pipelines used to validate the models.

\subsection{System Design}
\label{sec:System Design}

\begin{figure}[htb]
    \centering
    \includegraphics[width=7.5 cm]{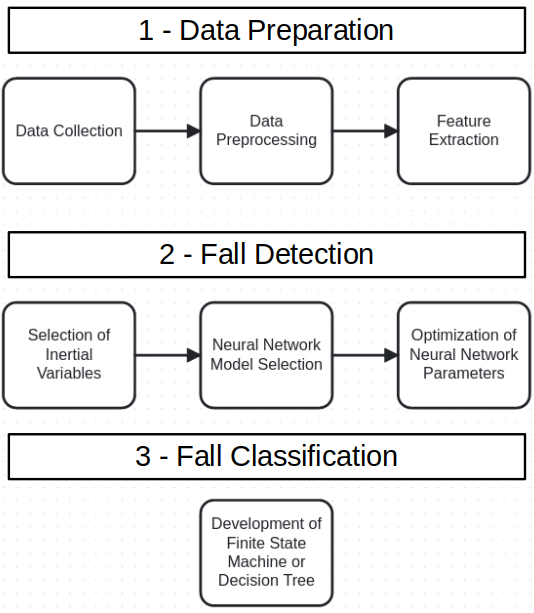}
    \caption{Outline of the Casualty Detection System.}
    \label{fig:outline}
\end{figure}

The Casualty Detection System is divided into three subsystems: 1 – Data Preparation, 2 – Fall Detection, and 3 – Fall Classification (Figure~\ref{fig:outline}). This paper addresses the research and results related to subsystems 1 and 2. In this section, we present the outline of the fall detection subsystem.

The fall detection problem can be addressed as a classification task using time series data from wearable devices sensors. In the proposed solution, a neural network is trained on a dataset containing fall incidents, daily activities, and military tasks. During operation, the neural network classifies each segment of data extracted using the sliding window technique, which partitions the time series into fixed-size windows, allowing each window to be independently classified by the neural network.

The neural network is specifically trained to identify the distinct patterns associated with a fall, marked by abrupt changes in both linear and angular acceleration of the soldier. Our objective is to equip the neural network with the ability to detect sudden deviations from regular movement patterns, even when they occur within a brief time frame. Mathematically, let $\{x_t\}_{t=1}^T$ denote the time series of acceleration data, where $x_t$ is the acceleration at time $t$. The neural network is trained on a dataset of falls and non-falls, composed of $N$ data windows. Each window is represented by a vector $y_i$, where $i = 1, 2, \ldots, N$. The goal of the fall detection algorithm is to classify each data window as a fall or non-fall. The neural network is trained to estimate a function $f(x_t)$ that maps the acceleration data to a probability of a fall for a window $i$, according to Equation~\ref{eq:1}, in which $\text{decision threshold}$ is a nonlinear function that maps input values to an output value of 0 or 1.

\begin{equation}
    \Pr(y_i = 1)= \text{decision threshold}(f(x_t)),
    \label{eq:1}
\end{equation}

In this paper, experiments were conducted to determine which sensors best captures the behavior of a fall: accelerometer, gyroscope, or a combination of both. These experiments will help identify which sensor should be used. Additionally, the decision threshold will be one of the parameters calibrated using Bayesian optimization technique to define the optimal threshold for fall classification.

Given the explained and observed in the data analysis section, that a fall activity has an approximate duration of five seconds, the fall detection algorithm to be used in real time during a military operation will function as follows:

\begin{enumerate}
    \item The wearable device, through its accelerometer and gyroscope sensors, will continuously record the data of the activity performed by the soldier.
    \item Concurrently with the recording of data by the wearable device, a sliding window will be initiated with a defined step of one second. When the time interval reaches a duration of five seconds, it will be considered a window, and the data will be processed. For better understanding, the example of the first three windows in an operation is cited: the first window starts at time 0s and ends at time 5s, the second window starts at time 1s and ends at time 6s, the third window starts at time 2s and ends at time 7s. With each step of one second, a new window will start and end after five seconds.
    \item The time series data of each window will be processed, that is, configured as data vectors and subsequently sent to be classified by the selected neural network.
\end{enumerate}

In this manner, the sliding window over time aims to find the exact pattern of a fall that was previously presented to the neural network during its training. If a fall starts at the second 2, it will last until the second 7. Thus, the start of the fall will occur in the first window (0 to 5 seconds), another part will occur in the second window (1 to 6 seconds), however, the complete pattern will be captured only by the third window (2 to 7 seconds). In other words, the algorithm will detect the fall only when the time interval of the third window ends.

\begin{figure}[htb]
    \centering
    \includegraphics[width=8cm]{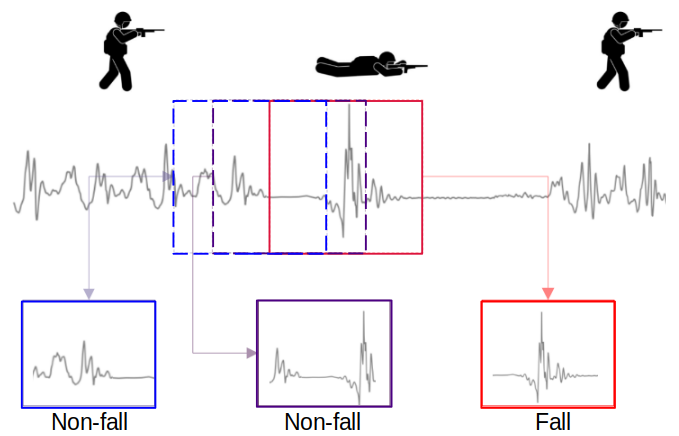}
    \caption{Fall Detection Algorithm Operation.}
    \label{fig:fda:operation}
\end{figure}

Therefore, the fall detection will not occur immediately but with a delay of a few seconds, depending on when the fall pattern is detected. Nevertheless, considering the ``golden hour'' and the ``platinum ten minutes'', this detection will be efficient, contributing to the situational awareness of the operation by the C2 and the quick mobilization of the rescue team. Figure~\ref{fig:fda:operation} displays the discussed scheme, where each window is represented by a dotted line. In the purple window, the beginning of the fall pattern is captured, which is only detected by the red window, where the complete fall pattern is captured.

\subsection{IPqM Fall Dataset}
\label{sec:IPqM Fall Dataset}

In this section, we provide an overview of the IPqM-FALL dataset, detailing the methodology for data collection and organization for fall detection. The aim was to create a comprehensive database for analyzing everyday activities, military operations, and falls, using inertial variables collected by wearable devices and smartphones. The following sections address the devices used, and the variables collected (Section~\ref{sec:Devices Used and Collected Variables}), the data collection methodology (Section~\ref{sec:Data Collection}), the voluntary participants (Section~\ref{sec:Voluntary Participants}), the collected Activities (Section~\ref{sec:Collected Activities}), and the organization of the dataset (Section~\ref{sec:Dataset Organization}). Each section offers insight into the methods employed to ensure the integrity and utility of the data for training and validating machine learning models for fall detection.

\subsubsection{Devices Used and Collected Variables}
\label{sec:Devices Used and Collected Variables}

The wearable devices selected by the ``Soldier of the Future'' project for data collection and subsequent use in operations were the Samsung Galaxy Watch 4 smartwatch and the LG Velvet smartphone. The Galaxy Watch 4 is a wearable device that not only connects to cell phones but is also capable of operating independently, recording biometric data even when not paired with a cell phone.

The Galaxy Watch 4 and the LG Velvet are equipped with a series of sensors, including an accelerometer, barometer, compass, geomagnetic sensor, gyroscope, among others.

The variables that were collected from the volunteer military's and will be observed in experiments using Neural Networks are the variables from the accelerometer and gyroscope sensors; they are:

\begin{itemize}
    \item Linear Acceleration: Obtained through the accelerometer sensor and measures the rate of change in the linear speed of a soldier. It provides information on the linear acceleration along the $x$, $y$, and $z$ axes.
    \item Angular Acceleration: Captured by the gyroscope sensor, it evaluates changes in the body's angular velocity over time. It provides information on the angular acceleration along the $x$, $y$, and $z$ axes.
\end{itemize}

\subsubsection{Data Collection}
\label{sec:Data Collection}

\begin{figure}[htb]
    \centering
    \includegraphics[width=6cm]{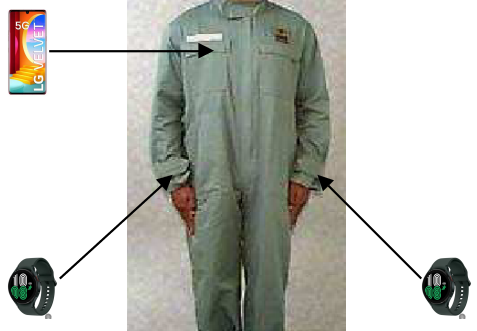}
    \caption{Positions of the devices used in the simulated activities.}
    \label{fig:position:devices}
\end{figure}

The data collection method was designed to capture data from multiple individuals simultaneously. Each volunteer military personnel used three devices during the activities: an Android smartphone in the chest pocket of the military uniform, along with two WearOS smartwatches, one on the left wrist and another on the right wrist (Figure~\ref{fig:position:devices}). Custom sampling apps were developed and installed on both the smartphones and smartwatches, along with an instructor app installed on an Android tablet used by the experiment coordinator. Concretely, two sampling apps were developed: one for mobile devices and another for wearable platforms.

The sampling application connects and receives commands from the instructor's application via HTTP using Server Sent Events (SSE), enabling the instructor's application to simultaneously send exercise start messages. This allows all connected devices to start sampling at the same time without the need for volunteer intervention. The start message includes a code to identify the exercise being started and the duration of the data collection period.

After collection, the application stores the data in a remote database management system. The stored data includes the identifier, timestamp, collected variables, location on the body (chest, right wrist, or left wrist), and the type of activity performed. 

The experiments were conducted in groups of 3 to 5 military personnel. The volunteers performed activities from three distinct categories: Activities of Daily Living, Military Operation Activities, and Fall Activities. All activities were supervised by a physical education specialist and a military personnel trained in military activities and firearm handling.During the activities, the sampling applications were programmed to vibrate as soon as the instructor initiated the command to start recording the activity, signaling to the volunteers the beginning of the execution.

The devices used for sampling were configured to capture data at the maximum rate using the SENSOR\_DELAY\_FASTEST setting, as recommended in the Android documentation~\cite{AndroidSensorManager}. As expected, sampling rates varied between the devices. The smartphone, due to its higher processing power, was able to achieve a higher sampling frequency. For linear acceleration, the phone had an average sampling rate of 221 Hz with a standard deviation of 23.4 Hz, while the smartwatches reached 92.6 Hz with a standard deviation of 1.25 Hz. For angular velocity, the smartphone showed an average sampling rate of 219.2 Hz with a standard deviation of 23.2 Hz, compared to the average of 91.7 Hz with a standard deviation of 1.25 Hz for the smartwatches. The collected data is stored along with its timestamp in milliseconds, allowing for subsampling or even interpolations to achieve more constant or lower sampling rates, according to the user's preference.

\subsubsection{Voluntary Participants}
\label{sec:Voluntary Participants}

To build the IPqM-Fall dataset, we collected data from 15 voluntary military personnel. To be included in the data collection experiment, the personnel had to be part of the daily armed service roster at IPqM, a service carried out on predetermined days when the personnel carry a firearm and their main function is to safeguard the military installations. Additionally, the volunteers had to have participated in rifle and pistol shooting training, be qualified for the daily armed service with both weapons, and have more than one year of service to ensure a minimum level of expertise. Therefore, volunteers who were unfit for armed service and military physical training were excluded.

Military personnel of both sexes and various ranks were selected as shown in Table~\ref{tab:personnel:info}. The selection of personnel from different ranks aims to ensure a diversity of experiences and skills, providing a more representative sample of the military population.

\begin{table}[htb] 
\caption{Age, rank, sex and anthropometric information of volunteer military personnel.}
\label{tab:personnel:info}
\begin{tabular}{cccccc}
\toprule
\textbf{ID}	& \textbf{Rank}	& \textbf{Gender}& \textbf{Age}& \textbf{Height (m)}& \textbf{Weight (Kg)}\\
\midrule
1& Seaman & Male & 23 & 1.63 & 68\\
2& Sergeant & Male & 34 & 1.96 & 88\\
3& Sergeant & Female & 36 & 1.64 & 68\\
4& Sergeant & Male & 30 & 1.71 & 70\\
5& Lieutenant & Male & 32 & 1.75 & 75\\
6& Sergeant & Female & 31 & 1.65 & 74\\
7& Lieutenant & Male & 34 & 1.74 & 80\\
8& Corporal & Male & 30 & 1.70 & 75\\
9& Lieutenant & Male & 38 & 1.70 & 95\\
10& Sergeant & Male & 37 & 1.71 & 70\\
11& Seaman & Male & 24 & 1.73 & 79\\
12& Corporal & Male & 27 & 1.71 & 56\\
13& Corporal & Female & 28 & 1.59 & 56\\
14& Lieutenant & Female & 29 & 1.64 & 58\\
15& Sergeant & Male & 33 & 1.71 & 87\\
\bottomrule
\end{tabular}
\end{table}

\begin{figure}[htb]
    \centering
    \includegraphics[width=7 cm]{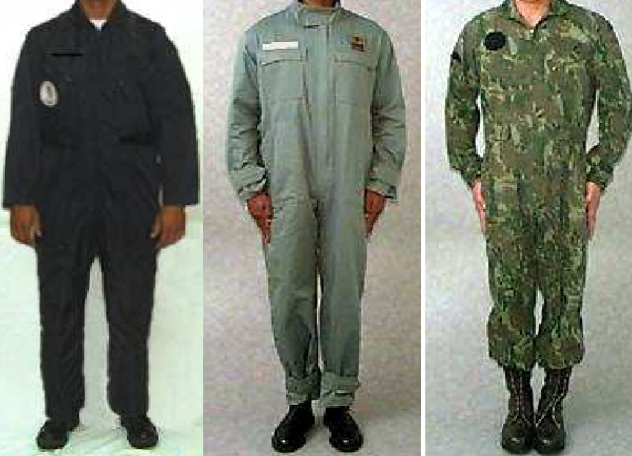}
    \caption{Uniforms OP1, AD, and OP2, respectively.}
    \label{fig:uniforms}
\end{figure}

The volunteer military personnel were instructed to report for the data collection exercise wearing uniforms OP1, OP2, or AD~\cite{rumb}, as shown in Figure~\ref{fig:uniforms}, to enable the use of the smartphone in the chest pocket.

The process of data collection with volunteer military personnel to compose the IPqM-Fall dataset was submitted to Plataforma Brasil (\url{https://conselho.saude.gov.br/plataforma-brasil-conep}), the unified Brazilian platform for human research records, and received approval from the Ethics Committee of Naval Hospital Marcílio Dias, under protocol CAAE 75570623.3.0000.5256. Written informed consent was obtained from all military volunteers involved in the data collection. 

\subsubsection{Collected Activities}
\label{sec:Collected Activities}

As mentioned earlier in Section~\ref{sec:Data Collection}, three distinct types of activities were collected: \textit{Activities of Daily Living} (ADL), which are basic tasks performed in everyday life; \textit{Military Operations Activities}, which involve tasks performed in military operations and service involving firearms; and \textit{Fall Activities}, which include situations involving loss of balance and contact of the body with the ground. These activities were selected to cover a wide range of situations experienced by a soldier, including those that might be confused with falling activities, with the aim of obtaining a diverse dataset for training neural networks or other machine learning techniques.

The ADL data collected were as follows: 1) standing, 2) walking, 3) running, 4) jumping, 5) sitting in a chair, 6) standing up from a chair, 7) walking uphill, 8) walking downhill, 9) running uphill, 10) running downhill, 11) walking upstairs, 12) walking downstairs, 13) jumping upstairs one step. The number of repetitions and the duration of these activities, along with instructions, are detailed in Table~\ref{tab:ADL}\footnote{The code numbering lacks $\text{ADL}_{9}$ and $\text{ADL}_{10}$ in Table~\ref{tab:ADL} because these simulations were not executed later, due to operational constraints. The same goes for $\text{FALL}_4$ in Table~\ref{tab:FA}.}.

\begin{table*}[htb] 
\centering
\caption{Activities of Daily Living.}
\label{tab:ADL}
\renewcommand{\arraystretch}{1.4}
\begin{tabular}{c p{2.8cm} p{0.4cm} p{1.29cm} p{6.9cm}}
\toprule
\textbf{Code} & \textbf{Activity} & \textbf{Reps} & \textbf{Duration} & \textbf{Description}\\
\midrule
$\text{ADL}_1$ & Standing & 1 & 3 min & Standing upright, performing subtle movements\\
$\text{ADL}_2$ & Walking & 1 & 3 min & Walking throughout the activity duration\\
$\text{ADL}_3$ & Running & 3 & 30 sec & Running throughout the activity duration\\ 
$\text{ADL}_4$ & Jumping & 3 & 30 sec & Standing, performing short jumps every 2 seconds\\
$\text{ADL}_5$ & Stair climbing & 3 & 10 sec & Walking upstairs\\
$\text{ADL}_6$ & Stair descending & 3 & 10 sec & Walking downstairs\\
$\text{ADL}_7$ & Sitting in chair & 6 & 5 sec & Starts standing, transitions to sitting in a chair\\
$\text{ADL}_8$ & Standing from chair & 6 & 5 sec & Starts sitting in a chair, transitions to standing\\ 
$\text{ADL}_{11}$ & Uphill walking & 3 & 30 sec & Walking uphill\\
$\text{ADL}_{12}$ & Downhill walking & 3 & 30 sec & Walking downhill\\
$\text{ADL}_{13}$ & Uphill running & 3 & 15 sec& Running uphill\\ 
$\text{ADL}_{14}$ & Downhill running & 3 & 15 sec & Running downhill\\ 
$\text{ADL}_{15}$ & Stair hopping & 3 & 10 sec & Climbing the stairs and jumping the space one step at a time\\
\bottomrule
\end{tabular}
\end{table*}

Activities $\text{ADL}_1$, $\text{ADL}_4$, $\text{ADL}_5$, $\text{ADL}_6$, $\text{ADL}_{11}$, $\text{ADL}_{12}$, $\text{ADL}_{13}$, and $\text{ADL}_{14}$ were also performed while carrying a rifle, considered as military operation activities with the same number and duration of repetitions indicated in Table~\ref{tab:ADL}. Activities $\text{ADL}_{13}$ and $\text{ADL}_{14}$ with a rifle are not conventional running, but rather speed engagement activities.

The Military Operation Activities (MO) were performed while carrying a rifle, except for the crawling activity. The collected activities were as follows: 1) Sweep with walking, 2) Sweep with quick engagement, 3) Transition from standing to kneeling shooting position, 4) Transition from walking to kneeling shooting position, 5) Transition from running to kneeling shooting position, 6) Transition from standing to prone shooting position, 7) Transition from walking to prone shooting position, 8) Transition from running to prone shooting position, 9) Crawl. The number and duration of the activities, as well as the descriptions, are detailed in Table~\ref{tab:AMO}.

\begin{table*}[htb] 
\caption{Activities of Military Operations.}
\label{tab:AMO}
\renewcommand{\arraystretch}{1.4}
\begin{tabular}{c l c c p{8.9cm}}
\toprule
\textbf{Code}& \textbf{Activity} & \textbf{Reps} & \textbf{Duration} & \textbf{Description}\\
\midrule
$\text{OM}_1$ & Sweep (Walking) & 1 & 3 min & Walking while sweeping (pointing the rifle at a coverage angle of approximately 120 degrees) \\
$\text{OM}_2$ & Sweep (Quick Engagement) & 3 & 45 sec & Quick Engaging while sweeping (pointing the rifle at a coverage angle of approximately 120 degrees) \\
$\text{OM}_3$ & Kneeling Shooting Position (Standing) & 3 & 10 sec & Starts standing, transitions to kneeling shooting position \\
$\text{OM}_4$ & Kneeling Shooting Position (Walking) & 3 & 1 min & Starts walking, transitions to kneeling shooting position \\
$\text{OM}_5$ & Kneeling Shooting Position (Running) & 3 & 45 sec & Starts running, transitions to kneeling shooting position \\
$\text{OM}_6$ & Prone Shooting Position (Standing) & 3 & 10 sec & Starts standing, transitions to prone shooting position \\
$\text{OM}_7$ & Prone Shooting Position (Walking) & 3 & 1 min & Starts walking, transitions to prone shooting position \\
$\text{OM}_8$ & Prone Shooting Position (Running) & 3 & 45 sec & Starts running, transitions to prone shooting position \\
$\text{OM}_9$ & Crawl  & 3 & 50 sec & Start in the crawling position and perform the Crawl activity for the duration of the activity. \\
\bottomrule
\end{tabular}
\end{table*}

The Falling Activities were conducted in two ways, with rifle and without rifle, as follows: 1) forward fall ending lying on the back, 2) forward fall ending lying face down, 3) backward fall, 4) right side fall, 5) left side fall. The side falls were alternated, with one group of volunteer military personnel performing the left side fall while another group performed the right side fall. The quantity, duration, and descriptions of the activities are detailed in Table~\ref{tab:FA}.

\begin{table*}[htb] 
\centering
\caption{Fall Activities. The rifle falls were performed with the same number of repetitions and duration.}
\label{tab:FA}
\begin{tabular}{c p{2.8cm} p{0.7cm} p{1.19cm} p{6.9cm}}
\toprule
\textbf{Code}& \textbf{Activity}& \textbf{Reps}& \textbf{Duration}& \textbf{Description}\\
\midrule
$\text{FALL}_1$ & Frontal Fall (Supine) & 6 & 10 sec & Falling forward from standing position to lying supine \\
$\text{FALL}_2$ & Frontal Fall (Prone) & 6 & 10 sec & Falling forward from standing position to lying prone \\
$\text{FALL}_3$ & Backward Fall & 6 & 10 sec & Falling backward from standing position to lying supine \\
$\text{FALL}_5$ & Lateral Fall - Right & 6 & 10 sec & Falling to the right from standing position \\
$\text{FALL}_6$ & Lateral Fall - Left & 6 & 10 sec & Falling to the left from standing position \\
\bottomrule
\end{tabular}
\end{table*}

For all activities involving the use of a rifle, aiming to minimize the risk of accidental discharge due to any errors in weapon unloading and to prevent damage to a real firearm during the activities, a decision was made to use an Airsoft M4 rifle as a substitute for the real firearm. The use of the Airsoft rifle will not compromise the results, as the chosen Airsoft M4 4003MG JG model (see Figure~\ref{fig:uniforms}) has similar dimensions and weight to a real M4 rifle. For instance, a Colt M4 Carbine rifle (Figure~\ref{fig:uniforms}) weighs 2.7 kg when unloaded and has a length of 81.28 cm, whereas the selected Airsoft model weighs 2.8 kg and has a total length of 78 cm.

\begin{figure}[htb]
\centering
\includegraphics[width=8cm]{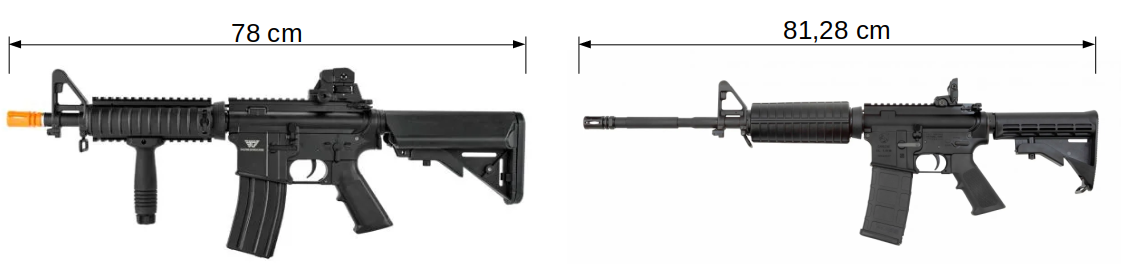}
\caption{Rifles airsoft M4 4003MG and Colt M4 Carbine, respectively.}
\label{fig:rifles}
\end{figure}

\subsubsection{Dataset Organization}
\label{sec:Dataset Organization}

The dataset is organized as follows: there are 15 directories corresponding to each volunteer military personnel. Within each directory, there is a file named README and three subdirectories named chest, left, and right, which contain data collected by the smartphone (on the chest) and by the smartwatches on the left and right wrists. The README file contains information about the volunteer military personnel, which was recorded in Table~\ref{tab:personnel:info}, including the number of activities and repetitions performed by the volunteer. Each subdirectory contains the two files in CSV format, as described below:

\begin{itemize}
    \item sampling.csv: Contains the activity ID, as well as the name of the exercise corresponding to the ID. Additionally, it indicates the sensor position on the body and the start and end timestamps of the activity recording. It has the \texttt{withRifle} field, which uses the number 0 to indicate that the activity was recorded without a rifle and the number 1 to indicate that it was recorded with a rifle. Through the activity ID, it is possible to identify the activities recorded in the acceleration.csv and angular\_speed.csv files. The \texttt{id} field of the sampling.csv file corresponds to the sampling field in the acceleration.csv and angular\_speed.csv files, functioning similarly to a foreign key.
    
    \item acceleration.csv and angular\_speed.csv. These files contain the sampling field, which identifies the recorded activity, the timestamp, and the $x$, $y$, and $z$ components of each observation.
\end{itemize}

Despite the significant contributions of existing fall detection datasets such as Mobiact~\cite{vavoulas2016mobiact}, Smartfall~\cite{mauldin2018smartfall}, Farseeing~\cite{klenk2016farseeing}, UMAFall~\cite{casilari2018umafall}, UP-Fall~\cite{martinez2019upfall}, OpenML~\cite{anguita2013public}, among others, none of them are entirely applicable to detecting a soldier's fall during an operation, as they do not collect military activities, especially activities involving firearms, such as engagement with a rifle, one of the activities mapped and collected in our experiments. After all, it is necessary to evaluate whether the rifle's weight, as well as the hand positions when carrying it, can result in a different acceleration pattern. Thus, the dataset published from the research presents an unprecedented approach that can be used by armed forces and police worldwide, and is also useful in fall research for the general population, by presenting in addition to armed activities, daily living activities and falls.

\subsection{Construction of the Fall Detection Neural Network Models}
\label{sec:Construction of the Fall Detection Neural Network Model}

This section discusses the steps involved in constructing the fall detection models, including data preprocessing, network architecture selection, and the training process. Additionally, it addresses key considerations such as feature extraction, model optimization, and the labeling process. 

\subsubsection{Choice of Neural Network Architecture}

In the context of integrating neural networks into devices like smartwatches and smartphones, it is essential to choose an architecture that balances effectiveness and simplicity, taking into account the computational resource limitations of these devices. In this regard, we chose to focus our experiments on the architecture of One-Dimensional Convolutional Neural Networks (CNN1D)~\cite{KIRANYAZ2021107398}. This architecture offers a suitable compromise between performance and efficiency, making it a viable choice for fall detection on portable devices such as smartphones and, especially, smartwatches.

In military operations, where mission durations are often unpredictable, the autonomy of fall detection devices becomes critical. This necessitates the use of low-complexity neural networks that can function efficiently and sustainably in such environments.

\subsubsection{Feature Extraction}
The accelerometer and gyroscope sensors of the wearable devices record measurements in the three components: $x$, $y$, and $z$ of \textit{linear acceleration} and \textit{angular acceleration}. In addition to these components, we extract the resultant force, a variable we call \textit{magnitude}. This variable combines the three components into a single measure of force, representing the total acceleration force applied to an object, considering the direction and intensity of each component of acceleration. Equation~\ref{eq:2} expresses the mathematical operation for calculating gravitational magnitude, where $a_x$, $a_y$ and $a_z$ represent the acceleration components in each axis.

\begin{linenomath}
    \begin{equation}
    R = \sqrt{a_x^2 + a_y^2 + a_z^2}
    \label{eq:2}
    \end{equation}
\end{linenomath}

We widely used the magnitude in data analysis and tested it in two sets of training pipelines. Our objective was to discover which combinations of variables were most efficient for fall detection.

\subsubsection{Data Analysis and Preparation}
\label{sec:Data Analysis}

To better understand the various patterns of activities collected in the IPqM-Fall dataset, we conducted an analysis of the data obtained from the military personnel. The analysis was carried out using the magnitude of linear acceleration from the accelerometer sensors of the wearable devices used. In this section, we present the data analysis and the subsequent processing performed based on this analysis.

\begin{figure}[htb]
    \centering
    \includegraphics[width=8cm]{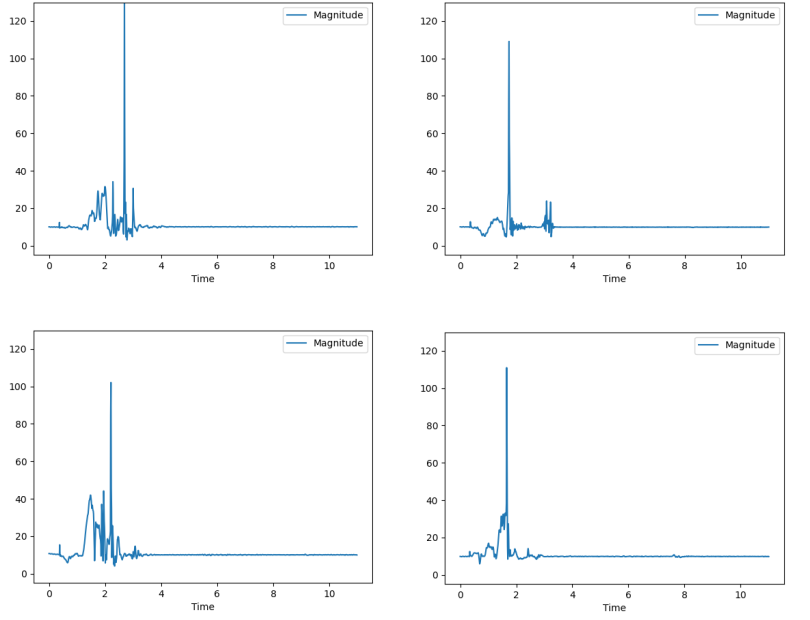}
    \caption{Fall activities recorded by the left smartwatch of ID 3: $\text{FALL}_1$, $\text{FALL}_2$, $\text{FALL}_3$, and $\text{FALL}_5$, respectively.
    \label{fig6}}
\end{figure}

\begin{figure}[htb]
    \centering
    \includegraphics[width=8cm]{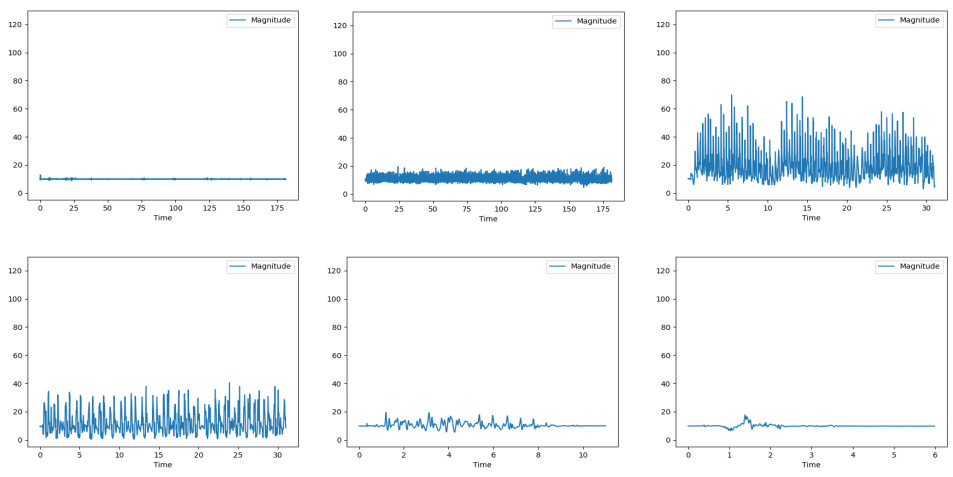}
    \caption{Fall activities recorded by the left smartwatch of ID 3: $\text{ADL}_1$, $\text{ADL}_2$, $\text{ADL}_3$, $\text{ADL}_4$, $\text{ADL}_5$ and $\text{ADL}_7$, respectively.
    \label{fig7}}
\end{figure}

The fall activities without a rifle have similar patterns (Figure~\ref{fig6}), showing a peak in the intensity of the acceleration magnitude when the body hits the ground. In most fall activities, this peak ranges from 70 to 130, often accompanied by one or two additional peaks above 40. On the other hand, activities of daily living (ADLs) do not exhibit similar patterns among themselves (Figure~\ref{fig7}), but they have distinct characteristics from fall activities. This allows for a binary class approach (Fall Activities and Non-Fall Activities), where one class represents all fall activities, with and without a rifle, and another class represents all daily life and military operations activities. The class labeled ``falls'' received the label 1, and the class labeled ``non-falls'' received the label 0.

\begin{figure}[htb]
    \centering
    \includegraphics[width=8cm]{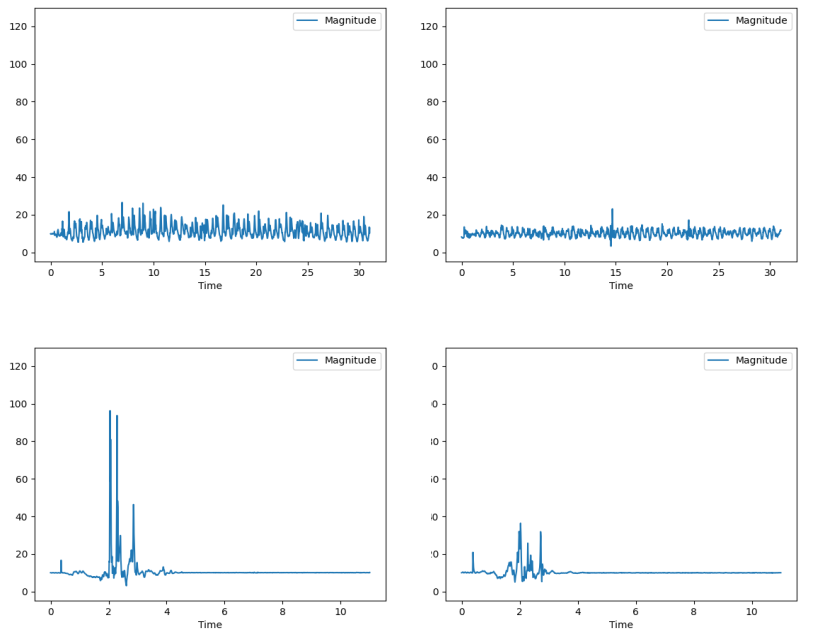}
    \caption{Fall activities recorded by the left smartwatch of ID 3: $\text{ADL}_{12}$, $\text{ADL}_{12}$\_withRifle, $\text{FALL}_3$ and $\text{FALL}_3$\_withRifle, respectively.
    \label{fig8}}
\end{figure}

There is a difference in the magnitude of activities that involve walking and running with and without a rifle ($\text{ADL}_2$, $\text{ADL}_3$, $\text{OM}_1$, and $\text{OM}_2$), with the magnitude of activities with a rifle being lower (Figure~\ref{fig8}). This is because walking with a rifle is an engagement walk, while running is a high-speed engagement activity, not exactly a walk or a run. Among the activities that involve falls with and without a rifle, the patterns are similar, but in many cases, the magnitude of falls with a rifle is lower (Figure~\ref{fig8}). Generally, the fall with or without the use of a rifle usually occurs in the first five seconds, followed by a stable pattern with an average acceleration magnitude of approximately 10 in the final five seconds, corresponding to the inactivity of the fallen body on the ground. Thus, the time window defined for fall detection is five seconds, corresponding, on average, to 450 observation points for the smartwatch and 1025 observation points for the smartphone. For both linear and angular acceleration, these averages are similar due to the data collection frequencies being close.

In this way, to generate the datasets in the time domain, data vectors containing 450 observations (right and left smartwatch) and 1025 observations (smartphone on the chest) were created for each fall activity, discarding the data related to the body in contact with the ground and keeping only the data related to the body's impact. After defining the size of the vectors for the falls, to maintain the same dimensionality in the input data for the tested neural networks, the size of the vectors corresponding to the other activities in IPqM-Fall was also fixed at 450 and 1025 observations.

Activities with data collected during trials of ten seconds or more, except for transition activities and $\text{ADL}_5$, $\text{ADL}_6$, and $\text{ADL}_1$5 activities related to going up and down stairs, exhibit a similar pattern in the magnitude variation over the activity duration. This allowed these activities to be segmented into more than one five-second set. For activities involving stairs, the set size was defined using only the first five seconds of the activity, discarding the rest of the collected data, as the staircase where the activities were collected was short, with only ten steps, and in most of the collected activities, after five seconds, the volunteer military personnel had already completed the ascent or descent of the stairs.

Due to these constant patterns and the fact that each activity was monitored for a different trial time, it was possible to divide them into multiple vectors, except for the already mentioned activities and activities with a duration of six seconds; for these, single data vectors of size 450 and 1025 were created, as was done for fall activities and activities using stairs.

In the transition activities to the shooting position ($\text{OM}_3$ to $\text{OM}_8$), such as the $\text{OM}_4$ activity, which is the transition from walking to the kneeling shooting position, the point of interest is the pattern caused by the state change, to verify with the neural network if, in any way, the activity is confused with a fall activity. Therefore, in these activities, only one five-second observation vector (480 and 1025) was selected and represents the moment of state change from one activity to another. In this way, the observation point where the highest magnitude peak occurs was identified, and the 240 observations before the peak and the 240 observations after the peak were selected to construct the data vector presented to the neural network, except for cases where it was not possible to distribute equally.

\begin{figure}[htb]
    \centering
    \includegraphics[width=8cm]{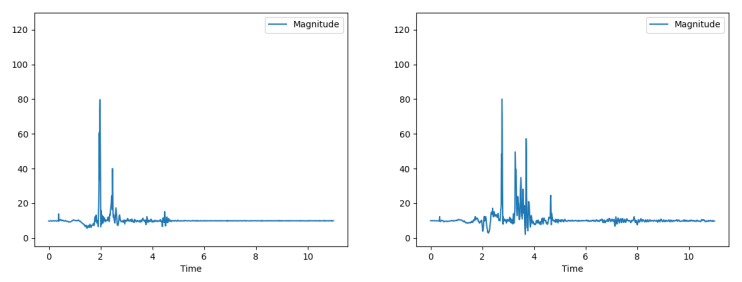}
    \caption{Activities recorded by the left smartwatch of ID 3: $\text{FALL}_3$ and $\text{OM}_6$, respectively.
    \label{fig9}}
\end{figure}

Fall activities and transition activities to the prone shooting position ($\text{OM}_6$, $\text{OM}_7$, and $\text{OM}_8$) exhibit similar patterns (Figure~\ref{fig9} because both involve the body hitting the ground, which can potentially confuse the neural network. However, as the first stage of the Soldier of the Future project aims to detect falls for subsequent classification if the fall occurred due to injuries or not, an adopted approach was to consider these transition activities as falls. Each experimental pipeline was executed with labels considering these activities as falls and, in another approach, considering them as non-fall activities.

The magnitude over time, evidenced in the analysis of fall activities, suggests that frequency domain analysis may also be useful. This analysis could, hypothetically, more precisely characterize fall activities and distinguish them from ADLs and military operations activities. Therefore, applying a Fourier Transform to the datasets and using this data to train a neural network could result in a more efficient classification than in the time domain. In light of this, a comparison was made between the fall classification results in the frequency and time domains.

For data generation in the frequency domain, a Fourier Transform was applied to each data vector of size 450 and 1025 observations of data vectors for the time domain, subtracting from the transform the DC level, which represents the zero-frequency component, the part of the signal that does not vary over time and represents the signal's average. In this way, the resulting data vectors were reduced to half the size of the observations of the vector in the time domain.

For the execution of all experimental pipelines (Section~\ref{sec:experimental:pipelines}), the pre-processed data for training the neural networks was divided into three sets: training, testing, and validation. The test and validation sets each correspond to 20\% of the samples, while the training set corresponds to 60\% of the samples. Bayesian optimization of the hyperparameters (Section~\ref{sec:Implementation of Neural Networks}) was performed using the training and validation data. The test set was not presented to the network during optimization or subsequent training, and was used only for making predictions and evaluating the models' generalization performance.

All steps of data set preparation for neural network training and the pipeline in Python code for reproducing data preparation and the experiments presented in this article are available in a GitHub repository\footnote{\url{https://github.com/AILAB-CEFET-RJ/falldetection}}.

\subsubsection{Experimental Pipelines and Labeling Approach}
\label{sec:experimental:pipelines}

In this section, we describe the experimental pipelines conducted for fall detection of soldiers using CONV1D neural networks. We designed the experiments to evaluate the effectiveness of different combinations of sensors (accelerometer and gyroscope) and the components of the inertial variables derived from these sensors in accurately detecting falls. The six training scenarios we tested were as follows:

\begin{enumerate}
     \item \textit{Using the magnitude of linear acceleration and the magnitude of angular acceleration individually}. In this scenario, the neural networks are trained separately using only the magnitude of linear acceleration and the magnitude of angular acceleration. In this scenario, the trained neural networks will be named as follows: Sc1Acc and Sc1Gyr for the CNN1D networks trained with the linear and angular acceleration data, respectively.
      
     \item \textit{Using the $x$, $y$, and $z$ components of linear acceleration and angular acceleration individually}. Here, the neural networks are trained separately, each using the three components ($x$, $y$, and $z$) of linear acceleration and angular acceleration. In this scenario, the trained neural networks will be named as follows: Sc2Acc and Sc2Gyr for the CNN1D networks trained with the linear and angular acceleration data, respectively.

     \item \textit{Using the magnitude of linear acceleration combined with the magnitude of angular acceleration}. In this experiment, the neural networks are trained with the magnitude of linear acceleration and the magnitude of angular acceleration simultaneously. In this scenario, the trained neural networks will be named as Sc3.

     \item \textit{Using the $x$, $y$, and $z$ components of linear acceleration combined with the $x$, $y$, and $z$ components of angular acceleration}. In this scenario, the neural networks are trained with the $x$, $y$, and $z$ components of linear acceleration and the $x$, $y$, and $z$ components of angular acceleration together. In this scenario, the trained neural networks will be named as Sc4.
    \end{enumerate}

The experiments were conducted with sensors positioned on the chest (smartphone) and on the right and left wrists (smartwatches). Testing sensors in different positions was performed to verify if there are variations in fall detection performance depending on the sensor's location. Additionally, testing both wrists is essential to determine if the position of the hands while holding a firearm (rifle) influences fall detection.

Tests for all mentioned scenarios were conducted with features in both the time domain and the frequency domain. To indicate this choice, the abbreviations for each scenario are accompanied by a ``T'' for the time domain or an ``F'' for the frequency domain. For example, we use Sc1AccT for the CNN1D network trained with linear acceleration data from scenario 1 in the time domain, and Sc1AccF for the same network trained in the frequency domain. Considering all variations (six scenarios, \textit{time domain} versus \textit{frequency domain}), the total number of resulting pipelines is twelve.

All pipelines were executed with two distinct labeling approaches, which we refer to as $\ell_1$ and $\ell_2$ (resulting in a total o twenty-four final fall-detection models). In both approaches, we worked with binary labels, namely, ``fall'' and ``not-fall'', as mentioned in Section~\ref{sec:Data Analysis}. However, $\ell_1$ considers the activities $\text{OM}_6$, $\text{OM}_7$, and $\text{OM}_8$ as fall activities. In contrast, in $\ell_2$, activities $\text{OM}_6$, $\text{OM}_7$, and $\text{OM}_8$ are considered non-falls. The resulting percentages of the positive class (``fall'') in labeling approaches $\ell_1$ and $\ell_2$ are $11.81\%$ and $9.82\%$, respectively.

\subsubsection{Hyperparameter search}
\label{sec:Implementation of Neural Networks}

To determine the most suitable hyperparameter configuration for the pre-processed data and the fall detection problem, and to ensure a fair comparison between the networks trained in the time and frequency domains, we used Bayesian optimization. This approach avoids empirical hyperparameter choices, ensuring that all networks in the experimental pipelines are efficiently optimized. The suggested values for the considered hyperparameters of a CNN1D architecture are presented in Table~\ref{tab:combined}. 

\begin{table}[htb] 
\caption{Suggested values for considered hyperparameters.}
\label{tab:combined}
\centering
\begin{tabular}{clc}
\toprule
\textbf{ID} & \textbf{Hyperparameter} & \textbf{Search Range}\\
\hline
$h_1$ & Number of Feature Maps$^{\mathrm{a}}$ & 8 to 600 \\
$h_2$ & Convolutional Filter Size (Kernel) & 2 to 6, by 1 \\
$h_3$ & Number of Convolutional Layers & 2 to 4, by 1 \\
$h_4$ & Number of Dense Layers & 1 to 3, by 1 \\
$h_5$ & Number of Neurons in Dense Layers$^{\mathrm{a}}$ & 60 to 320 \\
$h_6$ & Dropout Rate per Layer & 0.2 to 0.5, by 0.1 \\
$h_7$ & Learning Rate$^{\mathrm{b}}$ & 0.0001 to 0.01 \\
$h_8$ & Decision threshold & 0.5 to 0.9, by 0.1 \\
\bottomrule
\multicolumn{3}{p{220pt}}{$^{\mathrm{a}}$The search range was between the logs of the suggested values.}\\
\multicolumn{3}{p{220pt}}{$^{\mathrm{b}}$Complete list of values tried: 0.0001, 0.0003, 0.0006, 0.001, 0.003, 0.006, 0.01.}
\end{tabular}
\end{table}

Bayesian optimization was configured to maximize the \textit{Matthews Correlation Coefficient} (MCC), a statistical measure used to evaluate the quality of binary classifiers, especially efficient for imbalanced datasets. Values for MCC range from -1 to +1, where +1 indicates a perfect prediction, 0 indicates a random prediction, and -1 indicates a completely incorrect prediction. According to Equation~\ref{eq:mcc}, MCC takes into account the four elements of a binary classifier's confusion matrix: true positives (TP), true negatives (TN), false positives (FP), and false negatives (FN). MCC is considered a useful metric for evaluation in imbalanced classification problems~\cite{9440903,Chicco2020}.

\begin{equation}
\text{MCC} = \frac{\text{TP} \cdot \text{TN} - \text{FP} \cdot \text{FN}}{\sqrt{(\text{TP} + \text{FP})(\text{TP} + \text{FN})(\text{TN} + \text{FP})(\text{TN} + \text{FN})}}
\label{eq:mcc}
\end{equation}

Furthermore, the value for the \texttt{n\_trials} parameter of Optuna was set to 20. This parameter represents the number of iterations or experiments conducted during the search for the best hyperparameters. It is a measure of how many times the Bayesian optimization algorithm explores the search space to find the best combination of hyperparameters that maximizes the model's MCC. After execution, the ``trial'' with the best hyperparameter selection (best trial) was identified for each pipeline. Subsequently, the neural networks were trained 20 times using the ``best trial'', with the previously separated test sets. The results related to these trainings can be found in Section~\ref{sec:results}.

After performing Bayesian optimization for each pipeline, we obtained the hyperparameter configurations shown in Table~\ref{tab:opt:l1} (for labeling approach $\ell1$) and Table~\ref{tab:opt:l2} (for labeling approach $\ell2$). These hyperparameter combinations were used in the training sessions.

\begin{table}
    \caption{Optimized CNN1D architectures for different pipelines in labeling approach $\ell1$.}
    \label{tab:opt:l1}
    \begin{subtable}[t]{\linewidth}
        \centering
        \caption{Left Wrist}
            \begin{tabular}{ccccccccc}
            \hline
            \textbf{Pipeline} & $h_1$ & $h_2$ & $h_3$ & $h_4$ & $h_5$ & $h_6$ & $h_7$ & $h_8$ \\
            \hline
            Sc1AccT & 32 & 5 & 3 & 1 & 68 & 0.2 & 0.01 & 0.9 \\
            Sc1AccF & 466 & 3 & 3 & 3 & 198 & 0.3 & 0.0003 & 0.6 \\
            Sc1GyrF & 39 & 5 & 4 & 3 & 67 & 0.4 & 0.0003 & 0.5 \\
            Sc1GyrT & 344 & 3 & 2 & 3 & 284 & 0.3 & 0.0006 & 0.8 \\
            Sc2AccF & 301 & 3 & 4 & 2 & 170 & 0.2 & 1 & 0.6 \\
            Sc2AccT & 210 & 5 & 3 & 2 & 213 & 0.1 & 0.0006 & 0.5 \\
            Sc2GyrF & 78 & 6 & 4 & 2 & 85 & 0.2 & 0.0006 & 0.7 \\
            Sc2GyrT & 73 & 4 & 3 & 3 & 230 & 0.3 & 1 & 0.7 \\
            Sc3F & 136 & 5 & 3 & 2 & 182 & 0.2 & 0.0006 & 0.7 \\
            Sc3T & 100 & 3 & 3 & 3 & 180 & 0.1 & 0.0003 & 0.6 \\
            Sc4F & 52 & 3 & 2 & 3 & 78 & 0.4 & 0.0006 & 0.8 \\
            Sc4T & 300 & 2 & 2 & 1 & 194 & 0.4 & 3 & 0.5 \\
            \hline
            \end{tabular}
    \end{subtable}%
    
    \vspace{0.3cm}
    
    \begin{subtable}[t]{\linewidth}
        \caption{Right Wrist}
        \centering
        \begin{tabular}{ccccccccc}
            \hline
            \textbf{Pipeline} & $h_1$ & $h_2$ & $h_3$ & $h_4$ & $h_5$ & $h_6$ & $h_7$ & $h_8$ \\
            \hline
            Sc1AccT & 14 & 6 & 2 & 2 & 220 & 0.3 & 3 & 0.7 \\
            Sc1AccF & 72 & 3 & 2 & 2 & 63 & 0.1 & 0.0006 & 0.8 \\
            Sc1GyrF & 8 & 5 & 3 & 2 & 203 & 0.1 & 6 & 0.5 \\
            Sc1GyrT & 30 & 5 & 3 & 2 & 238 & 0.2 & 1 & 0.6 \\
            Sc2AccF & 20 & 3 & 2 & 3 & 214 & 0.4 & 1 & 0.7 \\
            Sc2AccT & 41 & 3 & 4 & 2 & 100 & 0.2 & 0.0006 & 0.5 \\
            Sc2GyrF & 44 & 3 & 2 & 2 & 89 & 0.2 & 1 & 0.9 \\
            Sc2GyrT & 9 & 4 & 4 & 3 & 66 & 0.1 & 1 & 0.5 \\
            Sc3F & 283 & 5 & 2 & 1 & 177 & 0.5 & 0.0003 & 0.6 \\
            Sc3T & 57 & 6 & 4 & 3 & 250 & 0.4 & 0.0006 & 0.7 \\
            Sc4F & 88 & 2 & 3 & 2 & 200 & 0.4 & 3 & 0.7 \\
            Sc4T & 40 & 2 & 2 & 2 & 173 & 0.1 & 0.0006 & 0.9 \\
            \hline
        \end{tabular}
    \end{subtable}%

    \vspace{0.3cm}
    
    \begin{subtable}[t]{\linewidth}
        \caption{Chest}
        \centering
        \begin{tabular}{ccccccccc}
            \hline
            \textbf{Pipeline} & $h_1$ & $h_2$ & $h_3$ & $h_4$ & $h_5$ & $h_6$ & $h_7$ & $h_8$ \\
            \hline
            Sc1AccT  & 37  & 3   & 4   & 2   & 107  & 0.2    & 3    & 0.6  \\ 
Sc1AccF  & 303 & 3   & 3   & 2   & 64   & 0.2    & 0.0001 & 0.6  \\ 
Sc1GyrF  & 83  & 2   & 2   & 3   & 62   & 0.5    & 1    & 0.5  \\ 
Sc1GyrT  & 293 & 3   & 2   & 3   & 188  & 0.2    & 0.0001 & 0.7  \\ 
Sc2AccF & 82  & 5   & 4   & 2   & 99   & 0.2    & 1    & 0.8  \\ 
Sc2AccT & 100 & 4   & 4   & 3   & 89   & 0.1    & 0.0003 & 0.9  \\ 
Sc2GyrF & 566 & 4   & 3   & 1   & 294  & 0.2    & 0.0003 & 0.5  \\ 
Sc2GyrT & 256 & 6   & 2   & 3   & 137  & 0.4    & 0.0006 & 0.7  \\
Sc\_3\_F     & 287 & 5   & 2   & 2   & 226  & 0.2    & 0.0003 & 0.7  \\ 
Sc3T     & 8   & 5   & 3   & 1   & 255  & 0.2    & 0.0006 & 0.6  \\ 
Sc4F     & 242 & 3   & 3   & 3   & 76   & 0.2    & 0.0006 & 0.7  \\ 
Sc4T     & 16  & 3   & 3   & 2   & 134  & 0.1    & 3    & 0.8  \\
\hline
        \end{tabular}
    \end{subtable}%
\end{table}

\begin{table}
    \caption{Optimized CNN1D architectures for different pipelines in labeling approach $\ell2$.}
    \label{tab:opt:l2}
    \begin{subtable}[t]{\linewidth}
        \caption{Left Wrist}
        \centering
        \begin{tabular}{ccccccccc}
            \hline
            \textbf{Pipeline} & $h_1$ & $h_2$ & $h_3$ & $h_4$ & $h_5$ & $h_6$ & $h_7$ & $h_8$ \\
            \hline
            Sc1AccT & 330 & 5 & 3 & 3 & 114 & 0.2 & 0.0006 & 0.7 \\
            Sc1AccF & 159 & 5 & 2 & 2 & 115 & 0.4 & 0.0006 & 0.9 \\
            Sc1GyrF & 239 & 5 & 2 & 2 & 64 & 0.1 & 1 & 0.8 \\
            Sc1GyrT & 220 & 6 & 4 & 1 & 74 & 0.2 & 1 & 0.8 \\
            Sc2AccF & 242 & 3 & 2 & 2 & 79 & 0.4 & 0.0003 & 0.5 \\
            Sc2AccT & 243 & 5 & 3 & 2 & 247 & 0.3 & 0.0006 & 0.9 \\
            Sc2GyrF & 73 & 4 & 3 & 3 & 230 & 0.3 & 1 & 0.7 \\
            Sc2GyrT & 27 & 3 & 2 & 1 & 61 & 0.4 & 1 & 0.7 \\
            Sc3F & 57 & 2 & 3 & 3 & 217 & 0.3 & 1 & 0.5 \\
            Sc3T & 8 & 6 & 2 & 1 & 298 & 0.2 & 1 & 0.8 \\
            Sc4F & 214 & 4 & 2 & 2 & 157 & 0.1 & 0.0006 & 0.9 \\
            Sc4T & 125 & 4 & 3 & 2 & 174 & 0.1 & 0.0003 & 0.6 \\       
            \hline
        \end{tabular}
    \end{subtable}%
    
    \vspace{0.3cm}
    
    \begin{subtable}[t]{\linewidth}
        \caption{Right Wrist}
        \centering
        \begin{tabular}{ccccccccc}
            \hline
            \textbf{Pipeline} & $h_1$ & $h_2$ & $h_3$ & $h_4$ & $h_5$ & $h_6$ & $h_7$ & $h_8$ \\
            \hline
            Sc1AccT & 253 & 6   & 3   & 2   & 206 & 0.2  & 0.0006 & 0.8  \\
        Sc1AccF & 217 & 5   & 4   & 3   & 290 & 0.2  & 0.0006 & 0.5  \\
        Sc1GyrF & 59  & 3   & 3   & 2   & 77  & 0.1  & 3     & 0.5  \\
        Sc1GyrT & 299 & 3   & 3   & 3   & 102 & 0.2  & 0.0006 & 0.8  \\
        Sc2AccF & 160 & 4   & 2   & 2   & 163 & 0.2  & 0.0003 & 0.6  \\
        Sc2AccT & 177 & 3   & 3   & 3   & 78  & 0.2  & 0.0003 & 0.8  \\
        Sc2GyrF & 109 & 4   & 3   & 1   & 144 & 0.2  & 0.0006 & 0.7  \\
        Sc2GyrT & 15  & 6   & 2   & 2   & 92  & 0.5  & 1     & 0.6  \\
        Sc3F & 550 & 5   & 2   & 3   & 92  & 0.2  & 1     & 0.5  \\
        Sc3T & 33  & 2   & 3   & 3   & 67  & 0.4  & 6     & 0.6  \\
        Sc4F & 88  & 6   & 2   & 3   & 64  & 0.5  & 0.0003 & 0.5  \\
        Sc4T & 9   & 6   & 2   & 1   & 74  & 0.1  & 0.01  & 0.9  \\
        \hline
        \end{tabular}
    \end{subtable}%
    
    \vspace{0.3cm}

    \begin{subtable}[t]{\linewidth}
        \caption{Chest}
        \centering
        \begin{tabular}{ccccccccc}
            \hline
            \textbf{Pipeline} & $h_1$ & $h_2$ & $h_3$ & $h_4$ & $h_5$ & $h_6$ & $h_7$ & $h_8$ \\
            \hline
            Sc1AccT & 12 & 4 & 4 & 1 & 76 & 0.1 & 3 & 0.7 \\
            Sc1AccF & 185 & 5 & 2 & 1 & 314 & 0.1 & 0.0001 & 0.6 \\
            Sc1GyrF & 151 & 5 & 2 & 3 & 103 & 0.3 & 0.0003 & 0.6 \\
            Sc1GyrT & 8 & 4 & 2 & 3 & 66 & 0.5 & 3 & 0.8 \\
            Sc2AccF & 39 & 6 & 2 & 2 & 151 & 0.3 & 0.0006 & 0.5 \\
            Sc2AccT & 222 & 2 & 3 & 3 & 155 & 0.2 & 0.0001 & 0.6 \\
            Sc2GyrF & 40 & 2 & 3 & 2 & 72 & 0.1 & 0.0003 & 0.5 \\
            Sc2GyrT & 598 & 4 & 3 & 2 & 99 & 0.4 & 1 & 0.5 \\
            Sc3F & 159 & 3 & 4 & 3 & 190 & 0.3 & 0.0003 & 0.6 \\
            Sc3T & 241 & 3 & 2 & 2 & 105 & 0.5 & 0.0006 & 0.7 \\
            Sc4F & 112 & 3 & 2 & 3 & 62 & 0.4 & 1 & 0.5 \\
            Sc4T & 38 & 5 & 4 & 2 & 171 & 0.2 & 0.0001 & 0.5 \\
            \hline
        \end{tabular}
    \end{subtable}%
\end{table}

\section{Results and Discussion}
\label{sec:results}

This section presents the conducted experiments and corresponding results. We performed experiments on the several variations of datasets (described in Section~\ref{sec:Construction of the Fall Detection Neural Network Model}). All the experiments were conducted on a server with two Nvidia GeForce GTX2080 GPUs, each one with 16GB memory. All the experiments were implemented in Python using the open-source framework TensorFlow (\url{https://www.tensorflow.org}) for model training. Bayesian hyperparameter optimization was performed with the help of the open-source Optuna framework (\url{https://optuna.org}). We begin by explaining the evaluation metrics. After that, we describe the results and a corresponding analysis.

The MCC metric (Eq.~\ref{eq:mcc}), which evaluates the balance between true positive, true negative, false positive, and false negative values, was used as the primary performance indicator. Additionally, false positives (FP) were analyzed to determine the reliability of the models, given that high FP rates compromise usability in real-world applications. This analysis was complemented with Sensitivity (SE, Eq.~\ref{eq:SE}), Specificity (ES, Eq.~\ref{eq:ES}), and Precision (PR, Eq.~\ref{eq:PR}) metrics for a more comprehensive overview. 

\begin{equation}
\label{eq:SE}
\text{SE} = \frac{\text{TP}}{\text{TP} + \text{FN}}
\end{equation}

\begin{equation}
\label{eq:ES}
\text{ES} = \frac{\text{TN}}{\text{TN} + \text{FP}}
\end{equation}

\begin{equation}
\label{eq:PR}
\text{PR} = \frac{\text{TP}}{\text{TP} + \text{FP}}
\end{equation}

The best results for labeling approaches ($\ell1$ and $\ell2$) are provided in Table~\ref{tab:bestresults_1} and in Table~\ref{tab:bestresults_2}, respectively. The number of true positives, true negatives, false positives, and false negatives obtained in the several experimental variations are also presented as separate tables in Appendix~\ref{sec:complete-results}.

\begin{table}
\caption{Best results for labeling approach $\ell_1$.} 
\label{tab:bestresults_1} 
    \begin{subtable}[t]{\linewidth}
        \caption{Left Wrist}
        \centering
        \begin{tabular}{crrrr}
            \toprule
            \centering Pipeline &
            \centering MCC &
            \centering SE &
            \centering ES &
            \centering PR \tabularnewline
            \midrule
            Sc1AccF & 0.8929 & 0.9905 & 0.8805 & 0.9821 \\
            Sc1AccT & 0.8886 & 0.9791 & 0.9371 & 0.9904 \\
            Sc1GyrF & 0.8001 & 0.9924 & 0.7233 & 0.9596 \\
            Sc1GyrT & 0.8690 & 0.9839 & 0.8805 & 0.9820 \\
            Sc2AccF & 0.9299 & 0.9962 & 0.9057 & 0.9859 \\
            Sc2AccT & 0.9677 & 0.9943 & 0.9811 & 0.9971 \\
            Sc2GyrF & 0.8862 & 0.9820 & 0.9182 & 0.9876 \\
            Sc2GyrT & 0.9466 & 0.9905 & 0.9686 & 0.9952 \\
            Sc3F    & 0.9323 & 0.9886 & 0.9560 & 0.9933 \\
            Sc3T    & 0.9186 & 0.9858 & 0.9497 & 0.9924 \\
            Sc4F    & 0.9466 & 0.9905 & 0.9686 & 0.9952 \\
            Sc4T    & 0.9640 & 0.9943 & 0.9748 & 0.9962 \\ 
            \hline
        \end{tabular}
    \end{subtable}%
    
    \vspace{0.3cm}
    
    \begin{subtable}[t]{\linewidth}
        \caption{Right Wrist}
        \centering
        \begin{tabular}{crrrr}
            \toprule
            \centering Pipeline &
            \centering MCC &
            \centering SE &
            \centering ES &
            \centering PR \tabularnewline
            \midrule
            Sc1AccF & 0.8760 & 0.9817 & 0.9067 & 0.9865 \\
            Sc1AccT & 0.8565 & 0.9759 & 0.9067 & 0.9864 \\
            Sc1GyrF & 0.7514 & 0.9759 & 0.7467 & 0.9639 \\
            Sc1GyrT & 0.8025 & 0.9894 & 0.7467 & 0.9644 \\
            Sc2AccF & 0.8608 & 0.9952 & 0.8000 & 0.9718 \\
            Sc2AccT & 0.9131 & 0.9875 & 0.9333 & 0.9903 \\
            Sc2GyrF & 0.8041 & 0.9865 & 0.7667 & 0.9670 \\
            Sc2GyrT & 0.8822 & 0.9923 & 0.8533 & 0.9791 \\
            Sc3F    & 0.8778 & 0.9798 & 0.9200 & 0.9883 \\
            Sc3T    & 0.8794 & 0.9827 & 0.9067 & 0.9865 \\
            Sc4F    & 0.8774 & 0.9981 & 0.8067 & 0.9728 \\
            Sc4T    & 0.8794 & 0.9827 & 0.9067 & 0.9865 \\
        \end{tabular}
    \end{subtable}%
    
    \vspace{0.3cm}
    
    \begin{subtable}[t]{\linewidth}
        \caption{Chest}
        \centering
        \begin{tabular}{crrrr}
            \toprule
            \centering Pipeline &
            \centering MCC &
            \centering SE &
            \centering ES &
            \centering PR \tabularnewline
            \midrule
            Sc1AccF & 0.8934 & 0.9972 & 0.8392 & 0.9788 \\
            Sc1AccT & 0.9202 & 0.9915 & 0.9231 & 0.9897 \\
            Sc1GyrF & 0.9085 & 0.9897 & 0.9161 & 0.9887 \\
            Sc1GyrT & 0.9212 & 0.9897 & 0.9371 & 0.9915 \\
            Sc2AccF & 0.9600 & 0.9972 & 0.9510 & 0.9935 \\
            Sc2AccT & 0.9843 & 0.9972 & 0.9930 & 0.9991 \\
            Sc2GyrF & 0.9114 & 0.9925 & 0.9021 & 0.9869 \\
            Sc2GyrT & 0.9490 & 0.9925 & 0.9650 & 0.9953 \\
            Sc3F    & 0.9273 & 0.9953 & 0.9091 & 0.9879 \\
            Sc3T    & 0.9254 & 0.9897 & 0.9441 & 0.9925 \\
            Sc4F    & 0.9841 & 1.0000 & 0.9720 & 0.9963 \\
            Sc4T    & 0.9881 & 0.9991 & 0.9860 & 0.9981 \\
        \end{tabular}
    \end{subtable}%
\end{table}

\begin{table}
\caption{Best results for labeling approach $\ell_2$.} 
\label{tab:bestresults_2}
    \begin{subtable}[t]{\linewidth}
        \caption{Left Wrist}
        \centering
        \begin{tabular}{crrrr}
            \toprule
            \centering Pipeline &
            \centering MCC &
            \centering SE &
            \centering ES &
            \centering PR \tabularnewline
            \midrule
            Sc1AccF & 0.8366 & 0.9797 & 0.8740 & 0.9852 \\ 
            Sc1AccT & 0.8817 & 0.9871 & 0.8976 & 0.9880 \\ 
            Sc1GyrF & 0.7991 & 0.9779 & 0.8268 & 0.9797 \\ 
            Sc1GyrT & 0.8672 & 0.9867 & 0.8616 & 0.9793 \\ 
            Sc2AccF & 0.9008 & 0.9871 & 0.9291 & 0.9917 \\ 
            Sc2AccT & 0.9606 & 0.9954 & 0.9685 & 0.9963 \\ 
            Sc2GyrF & 0.8465 & 0.9797 & 0.8898 & 0.9870 \\ 
            Sc2GyrT & 0.8865 & 0.9871 & 0.9055 & 0.9889 \\ 
            Sc3F    & 0.8979 & 0.9908 & 0.8976 & 0.9881 \\ 
            Sc3T    & 0.8770 & 0.9825 & 0.9213 & 0.9907 \\ 
            Sc4F    & 0.9181 & 0.9889 & 0.9449 & 0.9935 \\ 
            Sc4T    & 0.9648 & 0.9963 & 0.9685 & 0.9963 \\ 
        \end{tabular}
    \end{subtable}%
    
    \vspace{0.3cm}
    
    \begin{subtable}[t]{\linewidth}
        \caption{Right Wrist}
        \centering
        \begin{tabular}{crrrr}
            \toprule
            \centering Pipeline &
            \centering MCC &
            \centering SE &
            \centering ES &
            \centering PR \tabularnewline
            \midrule
            Sc1AccF & 0.8994 & 0.9916 & 0.8974& 0.9888 \\
            Sc1AccT & 0.8507 & 0.9832 & 0.8803& 0.9868 \\
            Sc1GyrF & 0.8076 & 0.9832 & 0.8120& 0.9795 \\
            Sc1GyrT & 0.8914 & 0.9888 & 0.9060& 0.9897 \\
            Sc2AccF & 0.9186 & 0.9935 & 0.9145& 0.9906 \\
            Sc2AccT & 0.9579 & 0.9944 & 0.9744& 0.9971 \\
            Sc2GyrF & 0.8759 & 0.9888 & 0.8803& 0.9869 \\
            Sc2GyrT & 0.8984 & 0.9869 & 0.9316& 0.9924 \\
            Sc3F    & 0.9392 & 0.9925 & 0.9573 & 0.9953 \\
            Sc3T    & 0.8820 & 0.9953 & 0.8376 & 0.9825 \\
            Sc4F    & 0.9575 & 0.9953 & 0.9658 & 0.9962 \\
            Sc4T    & 0.9169 & 0.9888 & 0.9487 & 0.9943 \\
        \end{tabular}
    \end{subtable}%
    
    \vspace{0.3cm}
    
    \begin{subtable}[t]{\linewidth}
        \caption{Chest}
        \centering
        \begin{tabular}{crrrr}
            \toprule
            \centering Pipeline &
            \centering MCC &
            \centering SE &
            \centering ES &
            \centering PR \tabularnewline
            \midrule
            Sc1AccF   & 0.9671 & 0.9954 & 0.9828 & 0.9982 \\
            Sc1AccT   & 0.9616 & 0.9973 & 0.9569 & 0.9954 \\
            Sc1GyrF   & 0.9582 & 0.9936 & 0.9828 & 0.9982 \\
            Sc1GyrT   & 0.9665 & 0.9973 & 0.9655 & 0.9963 \\
            Sc2AccF   & 0.9809 & 0.9982 & 0.9828 & 0.9982 \\
            Sc2AccT   & 0.9904 & 1.0000 & 0.9828 & 0.9982 \\
            Sc2GyrF   & 0.9858 & 0.9982 & 0.9914 & 0.9991 \\
            Sc2GyrT   & 0.8116 & 1.0000 & 0.6810 & 0.9672 \\
            Sc3F      & 0.9765 & 0.9963 & 0.9914 & 0.9991 \\
            Sc3T      & 0.9619 & 0.9963 & 0.9655 & 0.9963 \\
            Sc4F      & 0.9714 & 0.9973 & 0.9741 & 0.9973 \\
            Sc4T      & 0.9952 & 1.0000 & 0.9914 & 0.9991 \\
        \end{tabular}
    \end{subtable}%
\end{table}

Among the analyzed devices, the smartphone on the chest showed the best overall performance for both labeling approaches. In approach 1, pipeline Sc4T stood out with an MCC of 0.9881, sensitivity of 0.9991, specificity of 0.9860, and precision of 0.9981. In labeling approach $\ell_2$, pipeline Sc4T achieved an MCC of 0.9952, sensitivity of 1.0000, specificity of 0.9914, and precision of 0.9991. These values reinforce the high reliability and balance of the chest-mounted device.

Conversely, the devices on the left and right wrists showed inferior performance, with MCC generally below 0.96 in both labeling approaches. For instance, on the left wrist, the best MCC was 0.9648 (Sc4T, $\ell_2$), with a sensitivity of 0.9963 and specificity of 0.9685, while on the right wrist, it was 0.9575 (Sc4F, approach 1), with a sensitivity of 0.9981 and specificity of 0.8067.

Considering the results reported in Table~\ref{tab:bestresults_1} and in Table~\ref{tab:bestresults_2}, labeling approach $\ell_2$  consistently outperformed labeling approach $\ell_1$ in most pipelines, especially for the chest device. This reflects that labeling activities $\text{OM}_6$, $\text{OM}_7$, and $\text{OM}_8$ (corresponding to military transition activities to the lying position) as non-fall events improves the generalization capacity of the models.

Comparing the results across the three devices, we observe:

\begin{itemize}
    \item Left Wrist: The left wrist results, while consistent, were inferior to the chest in terms of MCC. The MCC ranged from 0.8001 to 0.9648 ($\ell_2$, Sc4T), with FP values generally exceeding 4, indicating a higher tendency for false alarms compared to the chest-mounted smartphone. Additionally, pipelines using time-domain data (T) showed higher average sensitivity than frequency-domain data (F), while specificity was more balanced. Precision was slightly higher in F pipelines.

    \item Right Wrist: The device on the right wrist showed similar performance to the left wrist, with a maximum MCC of 0.9575 ($\ell_1$, Sc4F) and FP values ranging from 2 to 38. Despite this, pipelines such as Sc4F ($\ell_1$) demonstrated competitiveness, achieving just 2 FPs. Here too, T pipelines had a slight advantage in sensitivity, while F pipelines excelled in precision and specificity.

    \item Chest: The chest-mounted smartphone led in overall performance. In particular, pipeline Sc4T ($\ell_2$) achieved MCC = 0.9952, FP = 1, sensitivity = 1.0000, and specificity = 0.9914. These results highlight the potential of the chest device as the preferred tool for fall detection due to its combination of high accuracy and low false positive rate. Additionally, T pipelines showed better sensitivity and specificity.
\end{itemize}

Comparing pipelines using time-domain data (T) with those using frequency-domain data (F), T pipelines generally exhibited superior MCC and lower FP rates. These results suggest that while F scenarios provide good precision, T pipeline offer a better balance between sensitivity and specificity, making them more suitable for fall detection.

Based on the analysis, the chest-mounted device combined with the labeling approach 2 demonstrated the best overall performance in terms of MCC, sensitivity, specificity, and false positives. Among the pipelines, Sc4T stood out as the most robust across all devices. This indicates that the most effective variable combinations for fall detection are the x, y, and z components of linear acceleration combined with the x, y, and z components of angular acceleration. Additionally, pipelines with time-domain data (T) proved superior to those with frequency-domain data (F), particularly for the chest-mounted smartphone. These results suggest that future implementations of fall detection systems should prioritize using chest-mounted devices with labeling similar to labeling approach 2 and time-domain configurations.

\section{Conclusion}
\label{sec:Conclusions}

This paper presented a machine learning-based method using one-dimensional convolutional neural networks ({CNN1D}) to detect soldier falls during military activities, aligning with the Brazilian Navy's project “Soldier of the Future”. The research aimed to automatically identify falls on the battlefield, contributing to the safety and operational support of combatants. Due to the lack of suitable public datasets, it was necessary to collect novel data from the Navy Research Institute, encompassing simulated falls, daily activities, and tactical military operations. The final model proved efficient in distinguishing falls from other activities.

The analysis explored both the time and frequency domains of the signals, with the Fourier Transform employed to extract frequency patterns. The {CNN1D} model, optimized through Bayesian methods for hyperparameter tuning, achieved the following results: MCC of 0.9952, sensitivity of 1.0000, and specificity of 0.9914 for the model trained with time-domain data collected from the soldier's chest, considering the $x$, $y$, and $z$ components of linear and angular acceleration.

The study's limitations include data collection in simulated environments and a relatively small number of participants (15 volunteers), which restricts the generalization of the results to broader and more varied situations. To address these limitations, the study proposes as future work data collection during real military training exercises, testing alternative architectures such as deep and hybrid neural networks, and exploring the use of ensemble techniques. The inclusion of data related to different weaponry and the analysis of the impact of gunfire and recoil on inertial variables could further enhance the model.

To support the research community on fall detection, we make data and source code publicly available on GitHub\footnote{\url{https://github.com/AILAB-CEFET-RJ/falldetection}} and Zenodo\footnote{\url{https://zenodo.org/records/12760391}}, ensuring the study's reproducibility and promoting continued research. Future advancements include the classification of falls as operational or incapacitating, broadening the system's applicability, and incorporating additional functionalities such as gunfire detection and support for ammunition control and combatant health monitoring. These developments could solidify the practical application of the model in a military context.


\appendix
\section{Complete results}
\label{sec:complete-results}

Table~\ref{tab:bestresults_l1} and Table~\ref{tab:bestresults_l2} present the number of true positives, true negatives, false positives, and false negatives obtained in the several experimental variations.

\begin{table}[ht]
\centering
\caption{Best results for labeling approach $\ell_1$.}
\label{tab:bestresults_l1}
\begin{tabular}{lcccc}
\hline
\textbf{Pipeline} & \textbf{TP} & \textbf{TN} & \textbf{FP} & \textbf{FN} \\
\hline
\multicolumn{5}{c}{\textbf{Left Wrist}} \\
\hline
Sc1AccF & 1043 & 140 & 19 & 10  \\
Sc1AccT & 1031 & 149 & 10 & 22  \\
Sc1GyrF & 1045 & 115 & 44 & 8   \\
Sc1GyrT & 0    & 140 & 19 & 17  \\
Sc2AccF & 1049 & 144 & 15 & 4   \\
Sc2AccT & 1047 & 156 & 3  & 6   \\
Sc2GyrF & 1034 & 146 & 13 & 19  \\
Sc2GyrT & 1043 & 154 & 5  & 10  \\
Sc3F    & 1041 & 152 & 7  & 12  \\
Sc3T    & 1038 & 151 & 8  & 15  \\
Sc4F    & 1043 & 154 & 5  & 10  \\
Sc4T    & 1047 & 155 & 4  & 6   \\ 
\hline
\multicolumn{5}{c}{\textbf{Right Wrist}} \\
\hline
Sc1AccF & 1020 & 136 & 19 & 14 \\
Sc1AccT & 1014 & 136 & 25 & 14 \\
Sc1GyrF & 1014 & 112 & 25 & 38 \\
Sc1GyrT & 1028 & 112 & 11 & 38 \\
Sc2AccF & 1034 & 120 & 5  & 30 \\
Sc2AccT & 1026 & 140 & 13 & 10 \\
Sc2GyrF & 1025 & 115 & 14 & 35 \\
Sc2GyrT & 1031 & 128 & 8  & 22 \\
Sc3F    & 1018 & 138 & 21 & 12 \\
Sc3T    & 1021 & 136 & 18 & 14 \\
Sc4F    & 1037 & 121 & 2  & 29 \\
Sc4T    & 1021 & 136 & 18 & 14 \\
\hline
\multicolumn{5}{c}{\textbf{Chest}} \\
\hline
Sc1AccF & 1062 & 120 & 23 & 3  \\
Sc1AccT & 1056 & 132 & 11 & 9  \\
Sc1GyrF & 1054 & 131 & 12 & 11 \\
Sc1GyrT & 1054 & 134 & 9  & 11 \\
Sc2AccF & 1062 & 136 & 7  & 3  \\
Sc2AccT & 1062 & 142 & 1  & 3  \\
Sc2GyrF & 1057 & 129 & 14 & 8  \\
Sc2GyrT & 1057 & 138 & 5  & 8  \\
Sc3F    & 1060 & 130 & 13 & 5  \\
Sc3T    & 1054 & 135 & 8  & 11 \\
Sc4F    & 1065 & 139 & 4  & 0  \\
Sc4T    & 1064 & 141 & 2  & 1  \\
\hline
\end{tabular}
\end{table}

\begin{table}[ht]
\centering
\caption{Best results for labeling approach $\ell_2$.}
\label{tab:bestresults_l2}
\begin{tabular}{lcccc}
\hline
\textbf{Pipeline} & \textbf{TP} & \textbf{TN} & \textbf{FP} & \textbf{FN} \\
\hline 
\multicolumn{5}{c}{\textbf{Left Wrist}}\\
\hline
Sc1AccF & 1063 & 111 & 22 & 16  \\
Sc1AccT & 1071 & 114 & 14 & 13  \\
Sc1GyrF & 1061 & 105 & 24 & 22  \\
Sc1GyrT & 1039 & 137 & 14 & 22  \\
Sc2AccF & 1071 & 118 & 14 & 9   \\
Sc2AccT & 1080 & 123 & 5  & 4   \\
Sc2GyrF & 1063 & 113 & 22 & 14  \\
Sc2GyrT & 1071 & 115 & 14 & 12  \\
Sc3F    & 1075 & 114 & 10 & 13  \\
Sc3T    & 1066 & 117 & 19 & 10  \\
Sc4F    & 1073 & 120 & 12 & 7   \\
Sc4T    & 1081 & 123 & 4  & 4   \\
\hline
\multicolumn{5}{c}{\textbf{Right Wrist}}\\
\hline
Sc1AccF & 1063 & 105 & 12 & 9   \\
Sc1AccT & 1054 & 103 & 14 & 18  \\
Sc1GyrF & 1054 & 95  & 22 & 18  \\
Sc1GyrT & 1060 & 106 & 11 & 12  \\
Sc2AccF & 1065 & 107 & 10 & 7   \\
Sc2AccT & 1066 & 114 & 3  & 6   \\
Sc2GyrF & 1060 & 103 & 14 & 12  \\
Sc2GyrT & 1058 & 109 & 8  & 14  \\
Sc3F    & 1064 & 112 & 5  & 8   \\
Sc3T    & 1067 & 98  & 19 & 5   \\
Sc4F    & 1067 & 113 & 4  & 5   \\
Sc4T    & 1060 & 111 & 6  & 12  \\
\hline
\multicolumn{5}{c}{\textbf{Chest}}\\
\hline
Sc1AccF & 1087 & 114 & 2  & 5   \\
Sc1AccT & 1089 & 111 & 5  & 3   \\
Sc1GyrF & 1085 & 114 & 2  & 7   \\
Sc1GyrT & 1089 & 112 & 4  & 3   \\
Sc2AccF & 1090 & 114 & 2  & 2   \\
Sc2AccT & 1092 & 114 & 2  & 0   \\
Sc2GyrF & 1090 & 115 & 1  & 2   \\
Sc2GyrT & 1092 & 79  & 37 & 0   \\
Sc3F    & 1088 & 115 & 1  & 4   \\
Sc3T    & 1088 & 112 & 4  & 4   \\
Sc4F    & 1089 & 113 & 3  & 3   \\
Sc4T    & 1092 & 115 & 1  & 0   \\
\hline
\end{tabular}
\end{table}

\bibliographystyle{unsrt}
\bibliography{References}

\end{document}